%% file: main.tex
\documentclass{article}

\PassOptionsToPackage{table}{xcolor}
\usepackage[dblblindworkshop, final]{neurips_2026}
\input{preamble}


\newcommand{\papertitle}{Goal Retrieval versus Action-Value Ordering:\\ Contrastive Critics under Best-of-\(K\) Selection}

\title{\papertitle}
\author{%
  Ayushman Singh\thanks{Work done independently outside Sesame.} \\
  Sesame AI \\
  \texttt{ayushman@sesame.com}
  \And
  Siddharth Aphale \\
  Stanford University \\
  \texttt{saphale@stanford.edu}
}
\hypersetup{pdfauthor={Ayushman Singh and Siddharth Aphale}}
\workshoptitle{From Pretrained Representations to Acting Agents (PTA)}

\begin{document}

\maketitle

\begin{abstract}
Goal-conditioned action generators can produce many plausible actions.
Best-of-\(K\) selection then uses a critic to rank those actions for the current
state and goal. Standard contrastive critics are trained and evaluated on a
different ranking: distinguishing future goals from unrelated negatives for a
fixed state--action pair. The unnormalized contrastive score is monotone in
behavior-policy value at the population optimum. Cosine-normalized scores lack
this guarantee, and finite trained networks may fail to recover the correct
ordering. We audit the resulting gap using shared state--goal queries and frozen
candidate pools, so critics differ only in their action ordering. Raw and cosine
critics reach retrieval AUCs of \(0.96\)--\(1.00\) across eight OGBench tasks.
Their progress ordering on navigation tasks is much weaker and sometimes
inverted. In the exact-value controlled problem, all three contrastive selectors
have higher mean regret than random selection. Their regret can increase with
\(K\). TD-Q stays close to the pool oracle. PointMaze is the only task with a
reliable learned value reference. TD-Q gives substantially better fixed-query
ordering there, though the differences in executed return are smaller and vary
by task. Matched controls suggest that graded value supervision accounts for
most of the improvement. Changing score bounds or model capacity alone has much
less effect. High retrieval performance does not establish whether a critic
ranks actions correctly for best-of-\(K\) selection. That requires a separate
action-ordering evaluation before deployment.
\end{abstract}

\input{sections/introduction}
\input{sections/related_work}
\input{sections/setup}
\input{sections/experimental_design}
\input{sections/controlled_selection}
\input{sections/score_semantics}
\input{sections/candidate_selection}
\input{sections/controls}
\input{sections/discussion}

\bibliographystyle{plainnat}
\bibliography{references}

\clearpage
\appendix
\input{supplement/supplement}

\end{document}

%% file: preamble.tex
\usepackage[utf8]{inputenc}
\usepackage[T1]{fontenc}
\usepackage{hyperref}
\usepackage{url}
\usepackage{amssymb,amsthm,mathtools,mathrsfs}
\usepackage{graphicx,subcaption}
\usepackage{booktabs,multirow}
\usepackage[expansion=false]{microtype}
\usepackage{nicefrac}
\usepackage{float}
\usepackage{afterpage}
\usepackage{wrapfig}
\usepackage{placeins}
\usepackage{xcolor}
\usepackage[capitalize,noabbrev]{cleveref}
\crefname{appendix}{Appendix}{Appendices}

\hypersetup{hidelinks}
\graphicspath{{figures/}}

\definecolor{heatcell}{HTML}{2E7D32}

%% file: sections/introduction.tex
\section{Introduction}
\label{sec:introduction}

Goal-conditioned action generators can return many plausible candidates. At
deployment, the question is which action makes the most progress toward the
goal. Standard contrastive training solves a different comparison: for a fixed
state--action anchor, it ranks compatible future goals above unrelated goals.
At the population optimum, the unnormalized contrastive density ratio orders
actions by the behavior policy's goal-conditioned value~\citep{eysenbach2022crl}.
A finite trained network may still get that ordering wrong. In this work, we test
whether trained bilinear critics retain the fixed-query action ordering required
by best-of-\(K\) selection.

Two actions may both make the intended goal easy to distinguish from an
unrelated negative, even when their values differ substantially. One might head
straight toward the goal while the other takes a detour. A retrieval test can
score both comparisons correctly. For best-of-\(K\) selection, the direct action
still has to outrank the detour. Larger candidate pools make this conflict more
consequential by concentrating selection on false positives in the score tail.

\begin{figure}[t]
  \centering
  \includegraphics[width=0.76\textwidth]{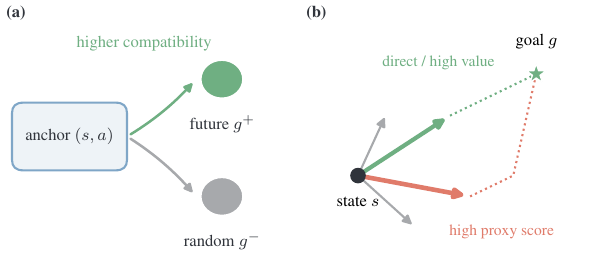}
  \caption{\textbf{From goal retrieval to action selection.} \textbf{(a)} Training ranks goals for a state--action anchor. \textbf{(b)} Deployment ranks actions for a fixed state and goal, where a detour can outscore direct progress. The two orderings agree at the population optimum.}
  \label{fig:teaser}
\end{figure}

Diffusion and flow-based action generators make large candidate sets practical,
and critic-guided methods use learned scalars to guide or rerank their samples~\citep{chi2023diffusionpolicy,wang2022diffusionql,lu2023qgpo,mao2024diffusiondice,dong2025expo,park2024fql}.
These methods generate and refine candidates differently. In every case, the
deployed scalar must rank actions within the same query. Standard contrastive
evaluation instead tests whether compatible future goals outrank unrelated
negatives~\citep{eysenbach2022crl,zheng2024stabilizing}; it therefore does not
test the action ordering consumed at deployment. \Cref{fig:teaser} illustrates
this mismatch, and \cref{fig:controlled-best-of-k} shows its exact-value
consequence: search improves the learned proxy while the selected action's value
falls.

We fix the state, goal, and candidate pool for each comparison. Three
contrastive readouts are then evaluated against TD-Q using exact values, a
reliability-gated value ensemble, and restored-state execution.
Across eight OGBench tasks, raw and cosine critics achieve retrieval AUCs of
\(0.96\)--\(1.00\) but weak or inverted navigation progress ordering. Every
contrastive selector has higher mean regret than random selection in the
controlled problem, and performance worsens as \(K\) grows. TD-Q stays near the
pool oracle.
PointMaze reproduces the fixed-query gap; executed-return differences are
smaller and task-dependent. Matched controls attribute most of the improvement
to graded value supervision.

\paragraph{Contributions.}
We examine why the population ordering guarantee may fail for finite or
cosine-normalized critics. The evaluation uses shared candidate pools to measure
action ordering under controlled conditions and connects those measurements to
execution. Our exact-value and OGBench experiments show that strong retrieval
can coexist with poor action ordering. Matched controls indicate that graded
value supervision accounts for much of the difference. These claims concern the
tested raw and cosine bilinear critics under candidate maximization, not every
contrastive objective or every closed-loop task. Action ordering should be
tested directly before a contrastive score is used to select among candidates.

%% file: sections/related_work.tex
\section{Related Work}

\paragraph{Contrastive and structured goal-conditioned critics.}
Contrastive RL represents state--action pairs and goals in a shared embedding space and trains their similarity using future goals and sampled negatives~\citep{eysenbach2022crl}. Subsequent work improves optimization and stability~\citep{zheng2024stabilizing}. Structured alternatives, including quasimetric and temporal-distance objectives, impose additional geometry that can encode graded reachability~\citep{wang2023quasimetric,myers2024temporal}. Our question is narrower: whether the deployed scalar learned by standard raw or cosine bilinear contrastive training orders the actions reached by candidate maximization. We do not compare all structured objectives and therefore do not claim that the observed gap applies to them.

\paragraph{Critic-guided candidate selection.}
Diffusion Policy exemplifies expressive action proposals that make large candidate sets practical~\citep{chi2023diffusionpolicy}. Critic-guided variants such as Diffusion-QL, QGPO, Diffusion-DICE, EXPO, and flow-policy methods use a learned critic or energy to guide or rerank actions and trajectories~\citep{wang2022diffusionql,lu2023qgpo,mao2024diffusiondice,dong2025expo,park2024fql}. We evaluate the deployed scalar directly, testing whether a standard bilinear contrastive score preserves action-value ordering within a fixed candidate pool.

\paragraph{Overoptimizing a learned proxy under search.}
Maximizing a learned scalar over a growing candidate set degrades the true objective once search reaches the proxy's error region. \citet{gao2023overoptimization} measure this for best-of-\(n\) reranking against a learned reward model and fit how gold-standard performance falls away from proxy performance as search intensifies. We study the same operation in goal-conditioned control, where the proxy is a contrastive critic and candidates come from an expressive action proposal. The goal-conditioned setting lets us examine ordering errors using an exact-value problem, a separately trained value reference, and executed rollouts. These comparisons help identify which ordering property fails.

\paragraph{Offline-RL extrapolation, calibration, and ranking.}
Offline RL has long studied exploitation of value errors outside the data distribution, motivating conservative and support-constrained methods~\citep{fujimoto2019bcq,kumar2020cql,kostrikov2021iql}. Calibration- and ranking-oriented approaches instead train the deployed scalar to preserve value levels or action order; RankQ is a recent example using explicit pairwise ordering constraints~\citep{nakamoto2023calql,choi2026rankq}. Our focus is on diagnosing the ordering problem. We show that strong goal retrieval does not establish the graded candidate ordering targeted by these value- and ranking-oriented objectives.

%% file: sections/setup.tex
\section{From Goal Retrieval to Candidate Ordering}
\label{sec:setup-problem}
\label{sec:why-fail}

A contrastive goal critic answers a compatibility question: \emph{given a
state--action pair, which goals are likely to be visited later?} For a
goal-conditioned state \(s\), action \(a\), and goal \(g\), a standard critic
represents the state--action pair and goal separately and assigns the score
\begin{equation}
    f_\theta(s,a,g)=\phi_\theta(s,a)^\top\psi_\theta(g).
    \label{eq:bilinear}
\end{equation}
The normalized variant uses cosine similarity,
\begin{equation}
    f_{\cos}(s,a,g)=
    \frac{\phi_\theta(s,a)^\top\psi_\theta(g)}
    {\lVert\phi_\theta(s,a)\rVert\lVert\psi_\theta(g)\rVert}.
    \label{eq:cos}
\end{equation}
We distinguish three types of ordering. \emph{Goal retrieval} compares candidate goals for a fixed \((s,a)\). \emph{Graded progress ordering} measures how scores track temporal progress across training-split triples from different queries. \emph{Fixed-query action ordering} compares candidate actions by value with \((s,g)\) held fixed. Best-of-\(K\) selection uses the last of these, while standard retrieval evaluation measures the first.

\paragraph{The population optimum is order-correct.}
These orderings do not conflict at the optimum. With future goals drawn from the geometric discount distribution of the behavior policy \(\pi\), the binary contrastive objective is minimized by
\begin{equation}
    f^\star(s,a,g) = \log\frac{p^{\pi}(g\mid s,a)}{p(g)} + c,
    \label{eq:density-ratio}
\end{equation}
the log density ratio between the discounted future-goal distribution and the marginal~\citep{eysenbach2022crl}. For fixed \((s,g)\), neither \(p(g)\) nor \(c\) varies with \(a\), so \(f^\star\) ranks actions by the discounted occupancy of \(g\), which is the goal-conditioned value of \(\pi\). An exactly optimized contrastive critic therefore supplies the ordering selection needs. Two qualifications drive the rest of the paper. \Cref{eq:density-ratio} describes the population optimum for the behavior policy. It does not tell us whether a finite trained critic preserves that ordering or how actions rank under an improved policy. It also characterizes an unconstrained logit: a cosine-constrained score cannot generally represent the same log-density ratio, so the corresponding ordering guarantee does not follow for \cref{eq:cos}.

\afterpage{%
  \begin{figure}[!t]
    \centering
    \includegraphics[width=\textwidth]{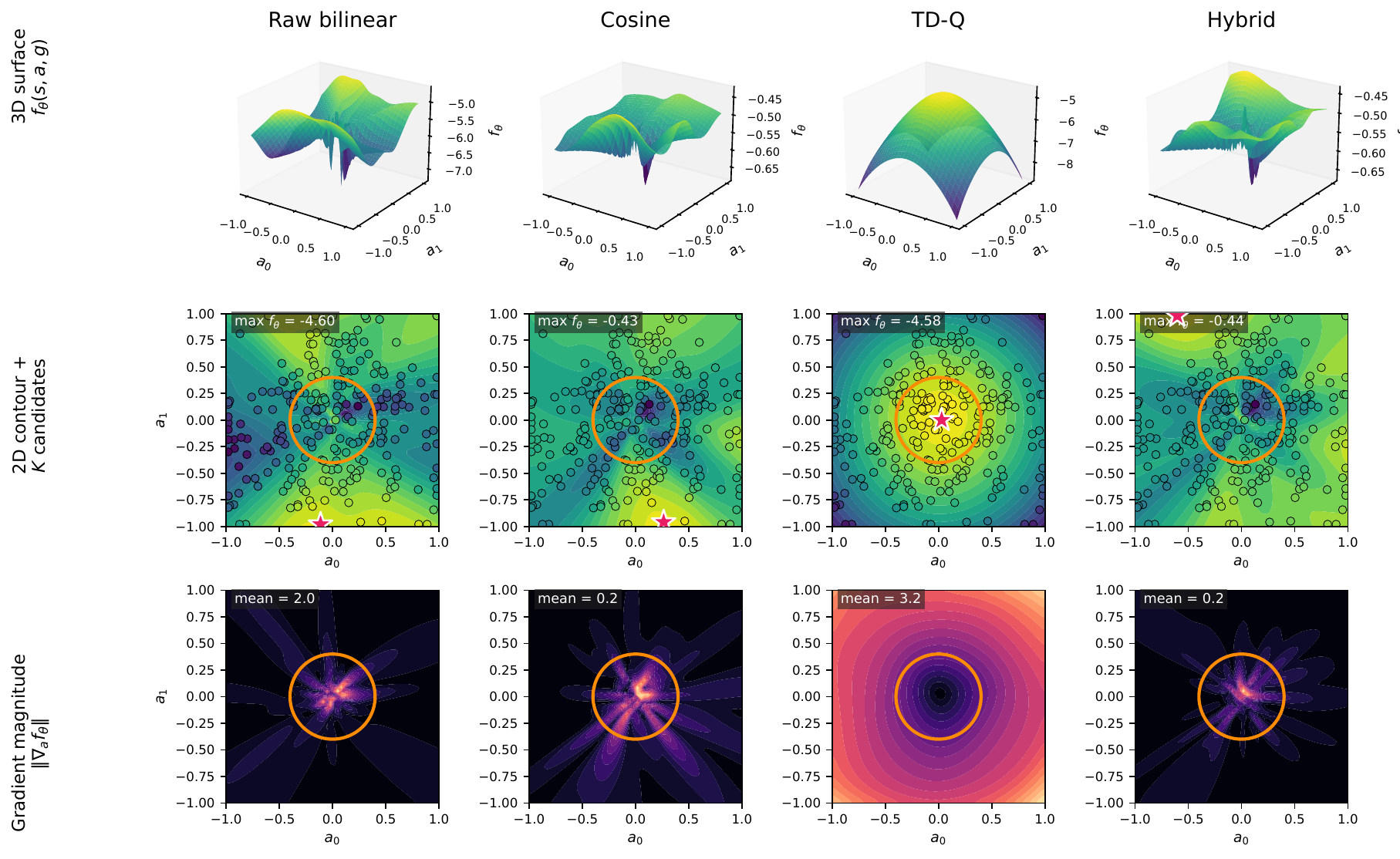}
    \caption{\textbf{Learned score geometry in the controlled two-dimensional problem.} Columns are the four deployed scores. Rows show the score surface, a contour with the shared candidates and the selected action $a_K^f$ (magenta star; orange support disk), and action-gradient magnitude. The contrastive readouts select outside support in this representative run, whereas TD-Q selects inside.}
    \label{fig:controlled-geometry}
    \label{fig:landscape}
  \end{figure}
  \afterpage{%
    \begin{figure}[!t]
      \centering
      \includegraphics[width=0.94\textwidth]{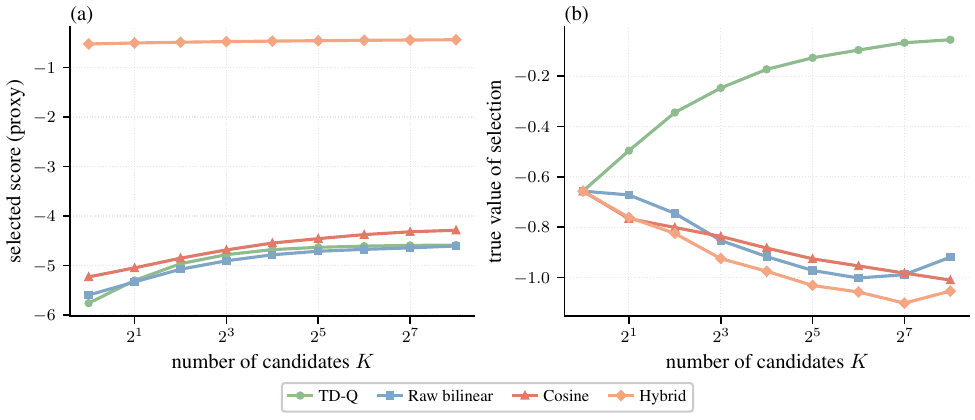}
      \caption{\textbf{Best-of-$K$ search in the controlled problem.} \textbf{(a) Proxy improves:} the selected critic score rises with $K$. \textbf{(b) Objective degrades:} the exact value of the selected action falls for each contrastive critic, whereas TD-Q retains low candidate regret (\cref{tab:controlled-selection}). Scores use each critic's native scale, so only within-critic trends are meaningful.}
      \label{fig:curse}
      \label{fig:controlled-best-of-k}
    \end{figure}
  }%
}

Let \(C_K=\{a_1,\ldots,a_K\}\) be actions sampled from a proposal \(\pi(\cdot\mid s,g)\). The critic selects
\begin{equation}
    a_K^f = \arg\max_{a_i\in C_K} f_\theta(s,a_i,g),
    \label{eq:rerank}
\end{equation}
whereas the candidate-pool oracle selects
\[
    a_K^Q = \arg\max_{a_i\in C_K} Q(s,a_i,g).
\]
Their candidate regret is
\begin{equation}
    \operatorname{Regret}_K
    = Q(s,a_K^Q,g)-Q(s,a_K^f,g).
    \label{eq:regret}
\end{equation}
This definition compares selectors on exactly the same candidate pool and therefore separates score ordering from proposal quality.

\paragraph{Why best-of-\(K\) search exposes high-score errors.}
Because \cref{eq:density-ratio} holds at the optimum, a selection failure reflects the estimation error of the trained critic, and best-of-\(K\) search is the operation that makes that error costly. This is the mechanism identified for best-of-\(n\) reranking against a learned reward model, where raising \(n\) improves the proxy score while the true objective eventually declines~\citep{gao2023overoptimization}. The counting argument is immediate. If the proposal assigns probability \(p>0\) to a low-value region whose critic scores exceed the scores of all alternatives, a \(K\)-candidate pool contains such an action with probability \(1-(1-p)^K\), which approaches one with \(K\). Candidate maximization therefore concentrates on whatever systematic false-positive region sits in the score tail.

The argument describes a possible selection failure and narrows down the kinds
of score errors that could cause it. These errors can arise from extrapolation
outside the dataset support, drifting embedding norms in an unnormalized
bilinear score, false alignment under a bounded score, or misordered actions
within support.
\cref{sec:controlled-selection,sec:candidate-selection,sec:controls} measure
which of these operate.

%% file: sections/experimental_design.tex
\section{Experimental Design}
\label{sec:experiments}
\label{sec:exp-setup}
\label{sec:protocols}

We begin with a controlled problem where exact values are known. We then examine
retrieval and progress scores across queries, evaluate selection on reliable
fixed-query candidate pools, and execute the selected actions from restored
simulator states. Each stage removes a different
source of ambiguity while preserving the same deployed-critic comparison;
matched controls then separate supervision from function class.

\paragraph{Critics.}
We evaluate four critics with distinct roles, all trained with the same data and training budget. Raw bilinear provides the standard unbounded contrastive score. Cosine tests whether normalization helps. Hybrid adds auxiliary TD supervision but still selects actions using cosine similarity; we include it as a diagnostic, not as a proposed method. TD-Q is a Bellman-trained joint-input twin-Q network that provides a value reference. \Cref{sec:controls} includes a matched \(2\times2\) comparison to account for differences in architecture and training objective.

\paragraph{Controlled problem.}
The controlled two-dimensional problem has transition \(s'=s+a\) and closed-form value
\begin{equation}
    Q^\star(s,a,g)=-\lVert s+a-g\rVert.
    \label{eq:toy-qstar}
\end{equation}
Dataset actions lie inside a disk of radius \(R\), while the candidate proposal samples both inside and outside that disk. Every evaluation goal is chosen so that the optimal action lies strictly inside support. This setting gives exact candidate regret and an unambiguous off-support label.

%% file: sections/controlled_selection.tex
\section{Controlled Candidate Selection with Exact Value}
\label{sec:controlled-selection}
\label{sec:controlled-case}

\begin{wraptable}{r}{0.46\textwidth}
  \centering
  \scriptsize
  \caption{\textbf{Controlled selection} at \(R=0.40\) and \(K=256\), with exact \(Q^\star\) and an in-support optimum. Random is exact over the frozen pools.}
  \label{tab:controlled-selection}
  \label{tab:toy-support}
  \renewcommand{\arraystretch}{0.92}
  \begin{tabular}{@{}lcc@{}}
    \toprule
    Critic & Regret & Off-support \\
    \midrule
    TD-Q & \(0.078{\pm}0.053\) & \(1{\pm}2\%\) \\
    Raw bilinear & \(0.978{\pm}0.106\) & \(94{\pm}8\%\) \\
    Cosine & \(0.644{\pm}0.346\) & \(60{\pm}37\%\) \\
    Hybrid & \(0.856{\pm}0.236\) & \(78{\pm}22\%\) \\
    \midrule
    Random pool action & \(0.570\) & \(71\%\) \\
    \bottomrule
  \end{tabular}
\end{wraptable}

Because the pools are frozen, a uniformly random pool action has exact regret
\(0.570\) and an exact off-support rate of \(71\%\), calibrating every critic
against chance. \Cref{fig:controlled-geometry} shows the learned score surfaces.
The three contrastive readouts place their selected maxima outside the support
disk in a representative run, despite the optimum lying strictly inside
support, whereas TD-Q selects inside.

The quantitative result is stable over seeds (\cref{tab:controlled-selection}).
TD-Q remains near the pool oracle. No contrastive readout has lower mean regret
than uniform random selection, and raw bilinear also has a higher mean
off-support rate than the random-pool base rate.

Hybrid's mean regret is \(0.856\pm0.236\), above random selection. Its
off-support rate of \(78\pm22\%\) is close to the random-pool baseline of
\(71\%\). Cosine likewise remains close to the random baseline
relative to its across-seed variability, with regret \(0.644\pm0.346\) and an
off-support rate of \(60\pm37\%\). Bounding the score reduces raw bilinear's
excess regret in this setting. The resulting action rankings remain unreliable.

\cref{fig:controlled-best-of-k} exhibits the upper-tail amplification mechanism from \cref{sec:setup-problem} directly. As \(K\) increases, the contrastive critics select actions with higher proxy scores but lower exact values. This pattern is consistent with score errors concentrated among the highest-scoring candidates. TD-Q instead tracks the pool oracle. An optimally trained contrastive score would order these candidates correctly (\cref{app:r-sweep}). The observed failure comes from estimation and extrapolation errors in the trained bilinear critic.

%% file: sections/score_semantics.tex
\section{OGBench: Strong Goal Retrieval, Weak Graded Progress Ordering}
\label{sec:score-semantics}
\label{sec:calibration}

\begin{figure}[t]
  \centering
  \includegraphics[width=0.94\textwidth]{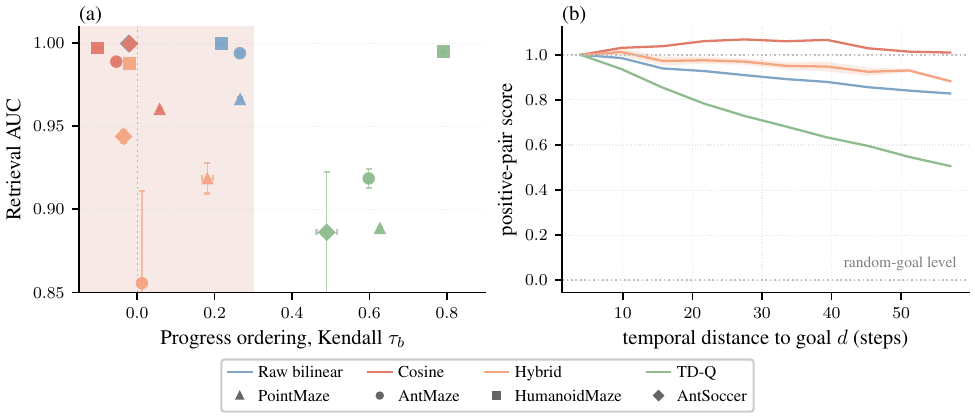}
  \caption{\textbf{Goal retrieval and progress ordering measure different properties on OGBench navigation.} \textbf{(a)} Retrieval AUC against Kendall \(\tau_b\) for the progress target \(\gamma^d\): contrastive critics occupy the high-retrieval, weak-ordering region, whereas TD-Q orders progress. \textbf{(b)} Positive-pair score over temporal distance. Both panels are cross-query diagnostics, not fixed-query ordering.}
  \label{fig:retrieval-vs-ordering}
  \label{fig:good-rankers}
\end{figure}

\paragraph{Cross-query score semantics.}
We next examine whether OGBench critics show the same separation between proxy scores and progress seen in the exact-value experiment. We evaluate eight OGBench tasks~\citep{park2024ogbench}: four navigation tasks (PointMaze, AntMaze, HumanoidMaze, and AntSoccer) and four manipulation-play tasks (Cube-single, Cube-double, Scene, and Puzzle). For a navigation triple whose future goal lies \(d\) steps ahead, \(\gamma^d\) is an in-support progress target, with larger values indicating nearer goals. Kendall \(\tau_b\) between score and \(\gamma^d\) measures graded progress ordering across queries (\cref{sec:setup-problem}). Undirected manipulation trajectories do not define an analogous progress target.

Raw and cosine critics achieve retrieval AUCs of \(0.96\)--\(1.00\) across all eight tasks, yet their scores track progress poorly on navigation tasks (\cref{fig:retrieval-vs-ordering}(a); Supplementary \cref{tab:retrieval-all}). Raw bilinear reaches \(\tau_b=0.22\)--\(0.27\) on the three maze tasks and has a slightly negative correlation on AntSoccer. Cosine correlations stay near or below zero. TD-Q reaches \(0.49\)--\(0.79\). Panel~(b) gives the same separation by temporal distance. The raw--cosine separation is consistent with the qualification in \cref{eq:density-ratio}: a cosine-constrained score need not represent the population-optimal log-density ratio. High retrieval AUC therefore provides no population-level guarantee of graded ordering for that readout.

\begin{figure}[!t]
  \centering
  \includegraphics[width=\textwidth]{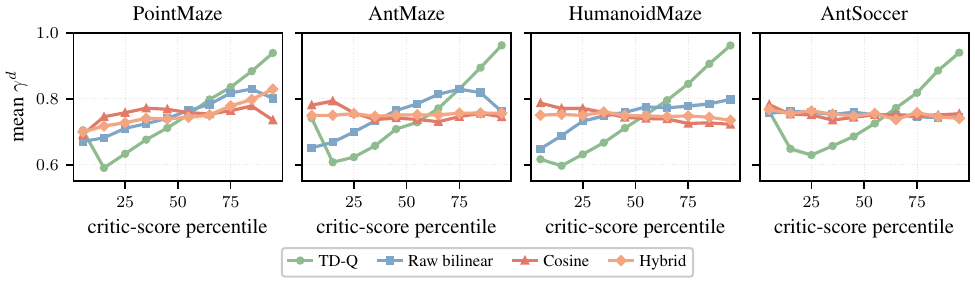}
  \caption{\textbf{Upper-tail progress ordering on OGBench navigation.}
    Held-out triples are binned by deployed-score decile and each bin reports
    mean $\gamma^d$. The upper-tail flattening or inversion is what best-of-$K$
    selection reads, and it motivates the separate fixed-query audit.}
  \label{fig:calibration}
\end{figure}

\paragraph{Robustness and implications for search.}
Nearest-state hard negatives preserve the ranking among critics, although absolute PointMaze retrieval AUC falls. Strong retrieval also extends to Cube-single, Cube-double, Scene, and Puzzle (Supplementary \cref{tab:retrieval-all}). TD-Q is a value-trained sanity check rather than independent evidence because its Bellman target is aligned with \(\gamma^d\); the contrastive critics do not receive this graded target, so their weak or inverted ordering is not implied by the evaluation construction.

\begin{figure}[!t]
  \centering
  \includegraphics[width=\textwidth]{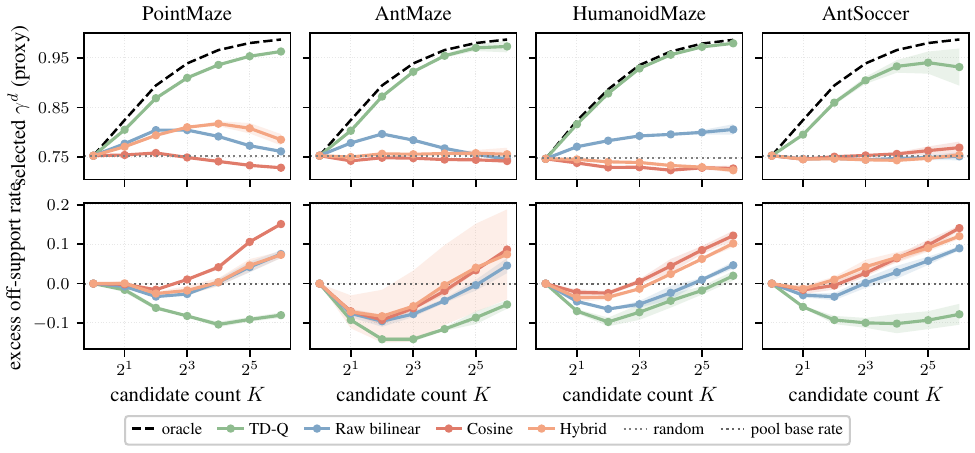}
  \caption{\textbf{Candidate-count diagnostics on OGBench navigation.}
    \textbf{Top:} Selected progress proxy $\gamma^d$ versus $K$, with the pool
    oracle dashed and the random baseline dotted. \textbf{Bottom:} Excess
    off-support selection over the pool base rate. Pools mix states and goals,
    so this diagnoses cross-query comparability rather than fixed-query
    ordering.}
  \label{fig:ogbench-selection}
\end{figure}

The separation becomes most consequential in the upper score tail, which is what best-of-\(K\) search consumes. The contrastive critics' progress scores flatten or turn downward in the upper deciles of \cref{fig:calibration}. TD-Q continues to track progress. The sampled navigation pools in \cref{fig:ogbench-selection} show a similar pattern as \(K\) grows, and the contrastive readouts increasingly select triples whose state--action neighborhoods have little local dataset support. These remain cross-query diagnostics, so \cref{sec:candidate-selection} next tests shared fixed-query action pools directly. Because selection depends only on score order, no strictly monotone post-hoc transformation can change these best-of-\(K\) argmaxes; \cref{app:additional} reports the full score-semantic, readout, retrieval, and hard-negative results.

%% file: sections/candidate_selection.tex
\section{Fixed-Query Selection and Executed Outcomes}
\label{sec:candidate-selection}
\label{sec:selection}

We now hold the state, goal, and candidate pool fixed to measure the action
ordering used at deployment. For each \(K=64\) pool,
we report within-pool Kendall \(\tau_b\) between the critic and reference
orderings and normalized regret (NRegret): zero matches the reference selector,
one matches the mean pool action, and values above one are worse than the pool
mean. A query is resolvable when the reference variation across its candidates
sufficiently exceeds disagreement among the reference models.

\begin{table}[t]
  \centering
  \scriptsize
  \caption{\textbf{Fixed-query PointMaze selection.} Results use 688 ensemble-resolvable queries.}
  \label{tab:fixed-query}
  \begin{tabular}{@{}lcc@{}}
    \toprule
    Critic & \shortstack{PointMaze\\$\tau_b@64$} & \shortstack{PointMaze\\NRegret@64} \\
    \midrule
    TD-Q & \(0.90{\pm}0.02\) & \(0.04{\pm}0.01\) \\
    Raw bilinear & \(0.41{\pm}0.01\) & \(0.66{\pm}0.02\) \\
    Cosine & \(0.23{\pm}0.01\) & \(1.15{\pm}0.02\) \\
    Hybrid & \(0.37{\pm}0.01\) & \(0.76{\pm}0.02\) \\
    \bottomrule
  \end{tabular}
\end{table}
\paragraph{Fixed-query protocol and PointMaze evidence.}
We freeze state--goal queries and candidate pools from a goal-conditioned diffusion behavior proposal, so every critic chooses from the same actions. We measure candidate regret against exact value or a separately trained three-member value ensemble on tasks that pass prespecified reliability gates. We execute each selected action from a restored simulator state, then follow the same continuation policy for every critic. Paired returns are measured against a random pool action and across selectors.\footnote{\label{fn:seeds}Reporting convention throughout. Critic-training seeds are \(\{42,43,44\}\); the controlled problem uses five seeds, and the restored-state audits use seed 42. Bands and \(\pm\) values are standard deviations. Rollout uncertainty is a paired query bootstrap.} \cref{app:critic-details} gives the complete protocol, reliability checks, sensitivity analyses, and rollout-precision calculations.

TD-Q closely tracks the reference-ensemble selector, whereas no contrastive
critic exceeds \(\tau_b=0.41\) (\cref{tab:fixed-query}). PointMaze is the only
task to pass the prespecified reliability gate; the primary estimand contains
688 resolvable queries. Because TD-Q and the reference both use value learning,
the exact-value experiment in \cref{sec:controlled-selection} provides the less
dependent comparison. Far-only and all-query sensitivities preserve the
separation (Supplementary \cref{tab:pointmaze-population-sensitivity}).

\begin{figure}[H]
  \centering
  \includegraphics[width=\textwidth]{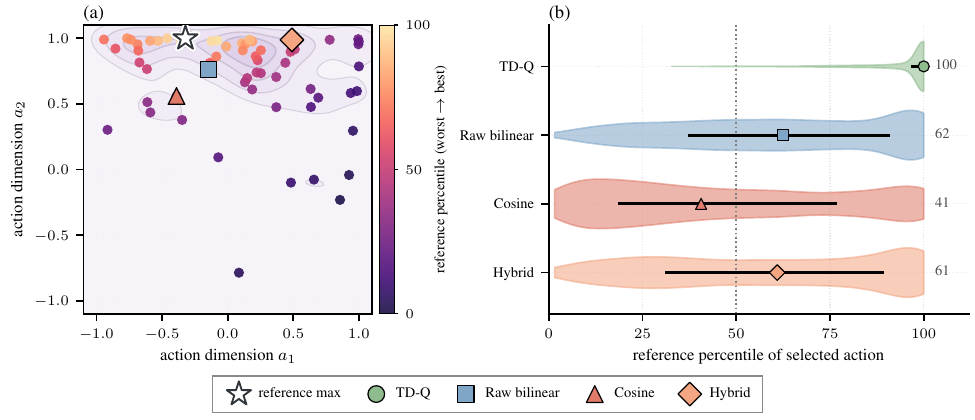}
  \caption{\textbf{What the PointMaze selectors choose at \(K=64\).} \textbf{(a)} A representative shared pool colored by reference-ensemble percentile. \textbf{(b)} Selected-action percentile across the 688 resolvable queries: TD-Q concentrates near the pool maximum; the contrastive selectors spread widely.}
  \label{fig:pointmaze-candidate-ranks}
\end{figure}

\begin{figure}[H]
  \centering
  \includegraphics[width=\textwidth]{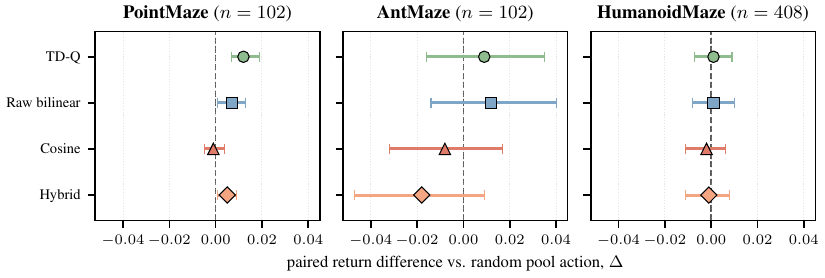}
  \caption{\textbf{Restored-state simulator audit.} Paired return differences relative to the same fixed random pool action, on a scale shared across tasks so the modest PointMaze effects and the wider AntMaze uncertainty are directly comparable. Each selector is compared with random rather than with the other selectors; Supplementary \cref{tab:paired-rollout} gives the pairwise contrasts.}
  \label{fig:rollout-effects}
\end{figure}

\paragraph{Executed outcomes and task boundaries.}
On PointMaze, TD-Q, raw bilinear, and hybrid improve mean discounted return over
random by \(0.012\), \(0.007\), and \(0.005\), whereas cosine is near zero
(\cref{fig:rollout-effects}). Paired comparisons favor TD-Q over cosine and
hybrid, but not raw bilinear (Supplementary \cref{tab:paired-rollout}). TD-Q
performs better than the bounded readouts in realized return. The difference is
smaller than its fixed-query ordering advantage might suggest after a
single-action intervention.

AntMaze and HumanoidMaze fail the reference-reliability gate, and every
selector's interval against random includes zero. HumanoidMaze meets the
rollout-precision criterion; AntMaze does not. AntSoccer lacks a usable proposal
and value reference and is therefore retained only in the cross-query score
analysis (\cref{app:critic-details}).

\FloatBarrier

%% file: sections/controls.tex
\section{What Causes the Ordering Gap?}
\label{sec:controls}
\label{sec:readout}
\label{sec:function-class}

\paragraph{Supervision versus architecture.}
\Cref{tab:objective-function-class} separates supervision from architecture in a
matched \(2\times2\) PointMaze comparison. Bellman supervision improves
\(\tau_b\) more than switching from a bilinear scalar to a joint-input MLP. Even
the weaker Bellman model outperforms the stronger contrastive model on ordering
and regret, though model capacity still matters.

\begin{table}[!htb]
  \centering
  \scriptsize
  \caption{\textbf{Objective versus function class} under the PointMaze fixed-query protocol. Each cell reports \(\tau_b@64\), followed by NRegret@64 in brackets.}
  \label{tab:objective-function-class}
  \label{tab:function-class}
  \setlength{\tabcolsep}{4pt}
  \renewcommand{\arraystretch}{0.92}
  \begin{tabular}{lcc}
    \toprule
    & Bilinear scalar & Joint-input MLP \\
    \midrule
    Contrastive & \(0.41{\pm}0.01\ [0.66{\pm}0.02]\) & \(0.46{\pm}0.00\ [0.56{\pm}0.00]\) \\
    TD / Bellman & \(0.70{\pm}0.01\ [0.33{\pm}0.00]\) & \(0.90{\pm}0.02\ [0.04{\pm}0.01]\) \\
    \bottomrule
  \end{tabular}
\end{table}

\paragraph{Score bounding versus value supervision.}
Raw-trained embeddings retain some ordering after cosine normalization.
Cosine-trained embeddings show weak ordering under either readout, and
state--action norm alone has almost no correlation with progress. In the
controlled problem, bounding reduces raw bilinear's off-support extrapolation
and regret (\cref{tab:controlled-selection}). PointMaze candidates are
in-support, so cosine instead removes score-scale information and worsens
ordering (\cref{tab:fixed-query}). Bounding reduces off-support
extrapolation in the controlled problem but worsens in-support ordering on
PointMaze. Neither embedding norm isolates the density ratio.

A two-head positive control shows that retrieval and graded value prediction are
compatible in this architecture. With separate encoders, the TD head reaches
cross-query \(\gamma^d\) Kendall correlations of \(0.59\)--\(0.79\) on the three
mazes, while the cosine head retains retrieval AUCs above \(0.96\)
(\cref{app:twohead}). The shared-backbone version is weaker on HumanoidMaze,
suggesting a separate value-supervised head rather than unrestricted parameter
sharing. The controls show that graded value supervision improves fixed-query
action ordering. Bounding or enlarging the contrastive score does not produce
the same improvement. These results leave open whether other forms of graded
supervision could work as well. The two-head result concerns cross-query progress ordering
rather than fixed-query action ordering.

\FloatBarrier

%% file: sections/discussion.tex
\section{Conclusion}
\label{sec:conclusion}

Contrastive training compares goals for a fixed state--action pair.
Best-of-\(K\) selection compares actions for a fixed state and goal. The two
orderings agree at the population optimum, though a trained critic may fail to
preserve that agreement. Our retrieval tests, controlled experiments, OGBench
candidate pools, and restored-state rollouts show that strong retrieval
performance does not guarantee the action ordering required by best-of-\(K\)
selection. Graded value supervision substantially improves ordering in the
matched controls. Score bounding alone does not. Before a contrastive
scalar is used for best-of-\(K\) selection, its ranking should be checked on
shared state--goal queries and tested through execution.

\paragraph{Limitations and future work.}
Our strongest evidence comes from the controlled problem and PointMaze; its
scope in higher-dimensional, longer-horizon tasks remains open. Future
experiments should include neutral ordering benchmarks and contrastive
objectives trained with graded value signals. We also need to test these
approaches with receding-horizon search, richer proposals, and larger domains.

%% file: supplement/supplement.tex
\section{Experimental Details}
\label[appendix]{app:critic-details}

\subsection{Critic settings and seeds}

All critics use the same offline data and training budget.
\Cref{tab:training-hyperparameters} lists the deployed scores and complete
training settings.

\begin{table}[H]
  \centering
  \scriptsize
  \caption{\textbf{Critic settings.} Goal-mix entries are
    current/trajectory/random; a dash denotes not applicable.}
  \label{tab:critics}
  \label{tab:training-hyperparameters}
  \setlength{\tabcolsep}{2.8pt}
  \renewcommand{\arraystretch}{0.94}
  \begin{tabular}{@{}p{0.24\linewidth}
    >{\centering\arraybackslash}p{0.16\linewidth}
    >{\centering\arraybackslash}p{0.16\linewidth}
    >{\centering\arraybackslash}p{0.16\linewidth}
    >{\centering\arraybackslash}p{0.16\linewidth}@{}}
    \toprule
    Setting & Raw bilinear & Cosine & TD-Q & Hybrid \\
    \midrule
    Deployed score & $\phi^\top\psi$ & $f_{\cos}$ & $Q_\omega$ & $f_{\cos}$ \\
    Bounded score & No & Yes & No & Yes \\
    Optimizer & Adam & Adam & Adam & Adam \\
    Learning rate & $3\times10^{-4}$ & $3\times10^{-4}$ & $3\times10^{-4}$ & $3\times10^{-4}$ \\
    Batch size & 1024 & 1024 & 1024 & 1024 \\
    Training updates & $10^6$ & $10^6$ & $10^6$ & $10^6$ \\
    Training seeds & $42,43,44$ & $42,43,44$ & $42,43,44$ & $42,43,44$ \\
    Hidden layers & 3 & 3 & 3 & 3 \\
    Units per hidden layer & 512 & 512 & 512 & 512 \\
    Layer normalization & Yes & Yes & Yes & Yes \\
    Critic ensemble size & 2 & 2 & 2 & 2 \\
    Embedding dimension & 512 & 512 & -- & 512 \\
    Embedding normalization & No & $\ell_2$ & -- & $\ell_2$ \\
    Critic objective & Binary BCE & Binary BCE & Bellman $+$ expectile & Binary BCE $+$ TD MSE \\
    Negatives per anchor & 1023 & 1023 & -- & 1023 \\
    Contrastive logit & $\phi^\top\psi/\sqrt{512}$ & $f_{\cos}/0.1$ & -- & $f_{\cos}/0.1$ \\
    Contrastive temperature & -- & 0.1 & -- & 0.1 \\
    Discount & 0.99 & 0.99 & 0.99 & 0.99 \\
    IQL expectile & -- & -- & 0.9 & -- \\
    Target update rate & -- & -- & 0.005 & -- \\
    BC coefficient & 3.0 & 3.0 & 3.0 & 3.0 \\
    Value-goal mix & $0/1/0$ & $0/1/0$ & $0.2/0.5/0.3$ & $0/1/0$ \\
    Auxiliary TD loss weight & -- & -- & -- & 1.0 \\
    Auxiliary TD start update & -- & -- & -- & 100,000 \\
    \bottomrule
  \end{tabular}
\end{table}

For the matched $2\times2$ control in \cref{tab:objective-function-class},
the Bellman-bilinear model uses the TD-Q target; the contrastive MLP matches
TD-Q's twin architecture and uses 127 negatives per anchor.

TD-Q is a separate GCIQL twin-Q network over $[s;g;a]$ and shares no parameters
with the contrastive encoders. Hybrid's auxiliary TD residual shapes the
representation, but selection uses only $f_{\cos}$. The PointMaze
objective/function-class control trains Bellman and contrastive objectives in
both the bilinear and joint-input MLP classes; the latter matches TD-Q's width,
layer normalization, and twin-network structure.

We use seeds $\{42,43,44\}$ and rerun only implementation, checkpoint, or
numerical failures, never based on performance. Paired bootstrap intervals
measure query and pool uncertainty; seed standard deviations measure training
variability.

\subsection{Cross-query and retrieval evaluation}

\textbf{Cross-query progress.}
For each navigation task, we sample 3,000 training-split triples $(s,a,g)$ from the
offline data. We sample $(s=o_t,a=a_t)$ and a future state
$g=o_{t+d}$ from the same trajectory, where
$d\sim\mathrm{Unif}\{1,\ldots,\min(60,\ell)\}$ and $\ell$ is the remaining
episode length. Each triple has the in-support progress target $\gamma^d$ with
$\gamma=0.99$. We report global Kendall $\tau_b$ between this target and each
deployed score. Since $\gamma^d$ is a strictly decreasing function of $d$, this
statistic equals the negated rank correlation with temporal distance, so the
discount sets the level in \cref{fig:calibration,fig:ogbench-selection}
without affecting any reported Kendall value.
Because this diagnostic pools triples with different states and goals, it
measures cross-query score semantics, not fixed-query action ordering.

\textbf{Retrieval.}
For all eight tasks, standard retrieval uses 3,000 analogous anchors with future
states as positives and goals from other trajectories as negatives. The
hard-negative variant instead chooses, from another trajectory, the goal whose
anchor state is nearest to the query anchor state. Its paired win rate is
reported as hard-negative AUC. We do not attach the navigation-only
$\gamma^d$ progress target to manipulation-play retrieval triples.

\subsection{Fixed-query evaluation}
\label{app:fixed-query-protocol}

\textbf{Candidate pools.}
We give every critic the same state--goal queries and candidate actions. We
freeze 1,500 queries from each
task's validation split: 500 each with future
goals at near ($d=1$--$10$), medium ($d=11$--$30$), and far
($d=31$--$60$) temporal distances. For every query we cache five independent
pools of $K=64$ actions sampled from a goal-conditioned DDPM proposal
$\pi(a\mid s,g)$. Samples are clipped only to the legal action bounds, and the
logged dataset action is stored as a tagged, nonselectable anchor. Pools are
generated before critic scoring; the same cached actions and nested prefixes
are shared across every critic and reference model.

\textbf{Reference gate.}
We require a learned judge's variation across candidates to exceed its internal
disagreement before using it for action-ordering claims.
The judge is an ensemble of three independently seeded GCIQL value functions
(seeds $\{7,8,9\}$), and candidate value is their mean. For each
query--replicate pool, let $S_q$ be the 90th-minus-10th percentile spread of
the ensemble mean and let $N_q$ be the median candidate standard error across
the three references. A pool is resolvable when $S_q>2N_q$; a query is
resolvable when a majority of its five pools are resolvable. We accept a task
for reference-judged conclusions only if at least 50\% and at least 100 queries
are resolvable in the full population or a prespecified distance stratum, and
the mean pairwise Spearman correlation among reference members is at least
0.70. This is a task-level gate. Once a task is accepted, the primary estimand
is the pooled set of individually resolvable queries across all three strata;
it is not restricted to the stratum that triggered acceptance. For PointMaze,
the far stratum triggers acceptance (280/500, 56.0\%) and the pooled resolvable
estimand contains 688/1,500 queries. Both rules were evaluated before critic
scores were inspected. \cref{tab:reference-reliability} shows why only
PointMaze is used for reference-judged action-ordering conclusions.

\begin{table}[H]
  \centering
  \caption{\textbf{Reference-ensemble reliability.} Best prespecified
    resolvable population and mean pairwise within-pool Spearman agreement.}
  \label{tab:reference-reliability}
  \begin{tabular}{lccc}
    \toprule
    Task & Best resolvable fraction & Agreement & Reference use \\
    \midrule
    PointMaze & $56.0\%$ (far) & $0.88$ & Accepted \\
    AntMaze & $27.2\%$ (medium) & $0.55$ & Excluded \\
    HumanoidMaze & $4.6\%$ (far) & $0.42$ & Excluded \\
    \bottomrule
  \end{tabular}
\end{table}

AntSoccer was scoped out before this protocol: it had no goal-conditioned
diffusion proposal and its roughly $5\%$-success GCIQL reference supplied
neither a usable value judge nor a precise rollout audit.

For accepted reference pools, the selector and candidate regret are
\begin{equation}
  \begin{aligned}
    a_K^f&=\arg\max_{a_i\in C_K}f(s,a_i,g), &
    a_K^Q&=\arg\max_{a_i\in C_K}\bar Q_{\mathrm{ref}}(s,a_i,g), \\
    \bar Q_C
    &=K^{-1}\sum_{a_i\in C_K}\bar Q_{\mathrm{ref}}(s,a_i,g), \\
    \mathrm{NRegret}_K
    &=\frac{\bar Q_{\mathrm{ref}}(s,a_K^Q,g)
      -\bar Q_{\mathrm{ref}}(s,a_K^f,g)}
      {\bar Q_{\mathrm{ref}}(s,a_K^Q,g)-\bar Q_C+10^{-9}}.
  \end{aligned}
\end{equation}
Values above one are possible when the selected action is worse than the mean
pool action. Kendall $\tau_b$ and NRegret are computed within each fixed
$(s,g)$ pool and then averaged over replicates and queries. Because deployed
TD-Q and the judge share the GCIQL value-learning family, TD-Q's low reference
regret is partly non-independent; the contrastive-versus-reference comparison
is the informative cross-objective result.

\begin{table}[H]
  \centering
  \small
  \caption{\textbf{PointMaze population sensitivity at $K=64$.}
    $\tau_b$/NRegret for the rollout-anchored deployed checkpoint.}
  \label{tab:pointmaze-population-sensitivity}
  \begin{tabular}{@{}lccc@{}}
    \toprule
    Critic & \shortstack{Pooled resolvable\\$n=688$} &
      \shortstack{Far resolvable\\$n=280$} &
      \shortstack{All queries\\$n=1{,}500$} \\
    \midrule
    TD-Q & $0.907/0.035$ & $0.918/0.033$ & $0.827/0.113$ \\
    Raw bilinear & $0.401/0.670$ & $0.377/0.718$ & $0.335/0.795$ \\
    Cosine & $0.232/1.146$ & $0.372/0.794$ & $0.174/1.279$ \\
    Hybrid & $0.368/0.782$ & $0.385/0.719$ & $0.273/0.914$ \\
    \bottomrule
  \end{tabular}
\end{table}

The TD-versus-contrastive separation persists in the pooled resolvable
population, the gate-passing far stratum, and all frozen queries.

\subsection{Restored-state rollout audit}

\textbf{Execution.}
The simulator audit executes one selected action from an exactly restored
validation state and then follows the same fixed, critic-independent
goal-conditioned DDPM continuation policy for every selector. PointMaze and
AntMaze states are restored from the observation representation; HumanoidMaze
is restored from the validation data's stored $qpos$ and $qvel$, which
reproduces the recorded observation to numerical precision. The horizon is 300
steps and the realized return is $\gamma^{T_g}$ if the goal is reached at step
$T_g$, and zero otherwise. Thus rollout differences are in discounted-return
units. We use five rollouts per action and common random numbers: the
continuation-policy random stream is identical for every selected action from
the same query and rollout replicate.

\textbf{Precision.}
PointMaze and AntMaze use the first 34 frozen queries per distance stratum
($102$ total), pool replicate zero. HumanoidMaze initially used the same set;
its prespecified frozen prefix was extended to 136 queries per stratum
($408$ total) after the initial interval failed the precision criterion. Each
reported $\Delta$ first averages rollouts within query, subtracts the same
query's fixed random-pool action, and then averages across queries. The 95\%
interval is a paired bootstrap over queries with 10,000 resamples; resamples
are shared across selectors.

The completed PointMaze audit yielded a TD-Q gain of approximately $0.012$;
we froze this prior observed scale as the smallest effect of interest,
$e_{\min}=0.012$, before examining HumanoidMaze intervals or extending its
query prefix. AntMaze predates this threshold and is assessed retrospectively.
A comparison is precise when its 95\% interval half-width is below
$e_{\min}$. PointMaze and HumanoidMaze meet this criterion; AntMaze's wider
intervals constitute a null finding, not a precise exclusion of an effect of
this size (\cref{tab:rollout-precision}).

\begin{table}[H]
  \centering
  \caption{\textbf{Rollout precision.} Ranges across TD-Q, raw, cosine, and
    hybrid; the precision threshold is a half-width below $0.012$.}
  \label{tab:rollout-precision}
  \begin{tabular}{lccc}
    \toprule
    Task & Queries & $\Delta$ range vs random & Half-width range \\
    \midrule
    PointMaze & 102 & $-0.001$ to $+0.012$ & $0.004$--$0.006$ \\
    AntMaze & 102 & $-0.018$ to $+0.012$ & $0.025$--$0.027$ \\
    HumanoidMaze & 408 & $-0.002$ to $+0.001$ & $0.008$--$0.010$ \\
    \bottomrule
  \end{tabular}
\end{table}

\textbf{Paired selector contrasts.}
The vs-random estimates do not establish differences between selectors.
Because every selector uses the same cached pool and continuation-policy random
stream, their per-query return differences are paired. \Cref{tab:paired-rollout}
uses the same estimator and 10,000-resample query bootstrap; identical selected
indices contribute exact zeros. Verdicts are stable across bootstrap seeds.

\begin{table}[H]
  \centering
  \caption{\textbf{Paired rollout contrasts.} Mean per-query discounted-return
    difference; positive favors TD-Q. Both tasks use $n=102$; HumanoidMaze
    records were unavailable, and comparisons are unadjusted for multiplicity.}
  \label{tab:paired-rollout}
  \begin{tabular}{llcc}
    \toprule
    Task & Contrast & $\Delta$ & 95\% interval \\
    \midrule
    \multirow{3}{*}{PointMaze}
      & TD-Q $-$ raw bilinear & $+0.005$ & Contains $0$; upper $+0.011$ \\
      & TD-Q $-$ cosine       & $+0.013$ & $[+0.006,+0.020]$ \\
      & TD-Q $-$ hybrid       & $+0.007$ & $[+0.001,+0.014]$ \\
    \midrule
    \multirow{3}{*}{AntMaze}
      & TD-Q $-$ raw bilinear & $-0.003$ & $[-0.031,+0.022]$ \\
      & TD-Q $-$ cosine       & $+0.017$ & $[-0.008,+0.041]$ \\
      & TD-Q $-$ hybrid       & $+0.027$ & $[+0.002,+0.053]$ \\
    \bottomrule
  \end{tabular}
\end{table}

On PointMaze, TD-Q has a large ordering advantage over raw bilinear ($\tau_b$ of
$0.90$ versus $0.41$). Their executed returns are not detectably different at the
measured precision, although TD-Q does outperform the two bounded readouts. A large one-step ordering gap
therefore need not produce a detectable realized-return gap after a one-action
intervention at this precision.
On AntMaze, only the TD-Q-versus-hybrid contrast excludes zero, and it is
marginal within a family of twelve comparisons.

\FloatBarrier

\section{Exact-Value Controlled Construction}
\label[appendix]{app:r-sweep}
\label{sec:running-example}

This construction makes both candidate value and dataset support known, so it
separates score-estimation and extrapolation error from reference-model error.
States, actions, and goals are two-dimensional,
$s,a,g\in\mathbb R^2$, with actions restricted to $[-1,1]^2$. The transition
is $s'=s+a$, the reward is $-\lVert s'-g\rVert$, and the exact one-step value is
\[
  Q^\star(s,a,g)=-\lVert s+a-g\rVert.
\]
The dataset action support is the disk
$\Omega=\{a:\lVert a\rVert\le R\}$. For each decision query we fix $s=0$ and
draw $g=\delta$ with $\lVert\delta\rVert\le0.6R$, so the optimal action
$a^\star=\delta$ lies strictly inside support.

\begin{table}[H]
  \centering
  \small
  \caption{\textbf{Controlled-construction settings.}}
  \label{tab:controlled-settings}
  \setlength{\tabcolsep}{5pt}
  \begin{tabular}{@{}llll@{}}
    \toprule
    Setting & Value & Setting & Value \\
    \midrule
    Dimension of $s,a,g$ & 2 & Dataset transitions & $20{,}000$ \\
    Action domain & $[-1,1]^2$ & Training updates & $50{,}000$ \\
    Main support radius & $R=0.40$ & Batch size & 256 \\
    Robustness radii & $\{0.25,0.40,0.55\}$ & Hidden layers & $2\times64$ \\
    Embedding dimension & 32 & Evaluation queries & 200 \\
    Candidates per pool & $K=256$ & Proposal std. & $1.2R$ per coordinate \\
    Discount & 0.99 & BC coefficient & 3.0 \\
    Main training seeds & 5 & Robustness seeds & 3 \\
    \bottomrule
  \end{tabular}
\end{table}

States and goals are drawn uniformly from $[-2,2]^2$. Dataset actions lie in
$\Omega$: half follow the normalized $g-s$ direction at a uniform radius, and
half are uniform in the disk, overrepresenting goal-reaching actions. The
contrastive positive is the recorded dataset goal, with in-batch goals as
negatives.

Evaluation goals and candidate pools are drawn once from a fixed generator seed
and reused across every critic and training seed. Candidate actions are sampled
independently from an isotropic Gaussian and clipped to the action bounds, so
pools contain both on- and off-support actions. At $R=0.40$, $71\%$ lie outside
the support disk, leaving roughly $75$ in-support candidates per pool, so a
chance baseline is needed to read the off-support column of
\cref{tab:controlled-selection}. Because the pools are fixed, that baseline
is exact rather than estimated: a uniformly random pool action has expected
regret $0.570$ and an off-support rate of $71\%$.

The learned critic selects $a_K^f$ and the exact pool oracle selects
$a_K^Q=\arg\max_{a_i\in C_K}Q^\star(s,a_i,g)$. We report exact candidate regret
$Q^\star(s,a_K^Q,g)-Q^\star(s,a_K^f,g)$ and the fraction of selections with
$\lVert a_K^f\rVert>R$. Across the robustness radii, contrastive selectors
remain predominantly off support while TD-Q remains near the exact pool oracle.

With deterministic transitions and dataset actions biased toward the goal, an
optimal contrastive score would order these candidates correctly. The observed
errors come from estimation and extrapolation in the trained bilinear model.

\FloatBarrier

\section{Supporting Retrieval and Score Diagnostics}
\label[appendix]{app:additional}

\begin{table}[H]
  \centering
  \small
  \caption{\textbf{Standard goal retrieval.} AUC for separating a future goal
    from a random goal on 3,000 training-split triples per task.}
  \label{tab:retrieval-all}
  \begin{tabular}{lcc}
    \toprule
    Task & Raw bilinear & Cosine \\
    \midrule
    PointMaze & $0.967{\pm}0.001$ & $0.961{\pm}0.000$ \\
    AntMaze & $0.994{\pm}0.000$ & $0.989{\pm}0.000$ \\
    HumanoidMaze & $1.000{\pm}0.000$ & $0.997{\pm}0.000$ \\
    AntSoccer & $1.000{\pm}0.000$ & $1.000{\pm}0.000$ \\
    \midrule
    Cube-single & $1.000{\pm}0.000$ & $1.000{\pm}0.000$ \\
    Cube-double & $1.000{\pm}0.000$ & $0.999{\pm}0.000$ \\
    Scene & $1.000{\pm}0.000$ & $1.000{\pm}0.000$ \\
    Puzzle & $0.999{\pm}0.000$ & $0.998{\pm}0.000$ \\
    \bottomrule
  \end{tabular}
\end{table}

\begin{table}[H]
  \centering
  \small
  \caption{\textbf{Hard-negative goal retrieval on navigation.} Hard negatives
    use the nearest anchor state from another trajectory; each task uses 3,000
    training-split triples.}
  \label{tab:retrieval}
  \begin{tabular}{llcc}
    \toprule
    Task & Critic & Standard AUC & Hard-negative AUC \\
    \midrule
    \multirow{4}{*}{PointMaze}
      & Raw bilinear & $0.967$ & $0.671$ \\
      & Cosine & $0.961$ & $0.642$ \\
      & TD-Q & $0.889$ & $0.514$ \\
      & Hybrid & $0.919$ & $0.611$ \\
    \midrule
    \multirow{4}{*}{AntMaze}
      & Raw bilinear & $0.994$ & $0.942$ \\
      & Cosine & $0.989$ & $0.906$ \\
      & TD-Q & $0.919$ & $0.833$ \\
      & Hybrid & $0.856$ & $0.748$ \\
    \midrule
    \multirow{4}{*}{HumanoidMaze}
      & Raw bilinear & $1.000$ & $0.995$ \\
      & Cosine & $0.997$ & $0.983$ \\
      & TD-Q & $0.995$ & $0.981$ \\
      & Hybrid & $0.988$ & $0.960$ \\
    \midrule
    \multirow{4}{*}{AntSoccer}
      & Raw bilinear & $1.000$ & $1.000$ \\
      & Cosine & $1.000$ & $0.999$ \\
      & TD-Q & $0.886$ & $0.861$ \\
      & Hybrid & $0.944$ & $0.873$ \\
    \bottomrule
  \end{tabular}
\end{table}

Standard AUC for the two contrastive critics spans $0.961$--$1.000$ across all
eight tasks. The hard-negative construction lowers absolute PointMaze AUC because its
two-dimensional observations make a nearby other-trajectory goal a difficult
alternative. Nevertheless, raw and cosine remain the strongest retrievers on
every navigation task. The retrieval/ordering separation therefore does not
depend on random negatives being intrinsically difficult.

\begin{table}[H]
  \centering
  \caption{\textbf{In-support graded progress ordering.} Kendall $\tau_b$
    between deployed score and $\gamma^d$, using 3,000 training-split navigation
    triples per task.}
  \label{tab:main}
  \begin{tabular}{lcccc}
    \toprule
    Critic & PointMaze & AntMaze & HumanoidMaze & AntSoccer \\
    \midrule
    Raw bilinear & $+0.266{\pm}0.005$ & $+0.265{\pm}0.005$ & $+0.218{\pm}0.007$ & $-0.022{\pm}0.002$ \\
    Cosine & $+0.058{\pm}0.001$ & $-0.054{\pm}0.005$ & $-0.102{\pm}0.007$ & $-0.020{\pm}0.002$ \\
    TD-Q & $+0.627{\pm}0.002$ & $+0.598{\pm}0.002$ & $+0.791{\pm}0.002$ & $+0.490{\pm}0.027$ \\
    Hybrid & $+0.181{\pm}0.015$ & $+0.012{\pm}0.005$ & $-0.020{\pm}0.004$ & $-0.035{\pm}0.011$ \\
    \bottomrule
  \end{tabular}
\end{table}

\begin{table}[H]
  \centering
  \caption{\textbf{Training/readout decomposition.} Kendall $\tau_b$ against
    $\gamma^d$ under alternative readouts.}
  \label{tab:readout}
  \begin{tabular}{lccc}
    \toprule
    Training / readout & PointMaze & AntMaze & HumanoidMaze \\
    \midrule
    Raw trained, raw $\phi^\top\psi$ & $+0.266{\pm}0.005$ & $+0.265{\pm}0.006$ & $+0.217{\pm}0.007$ \\
    Raw trained, cosine readout & $+0.158{\pm}0.004$ & $+0.215{\pm}0.004$ & $+0.219{\pm}0.007$ \\
    Raw trained, $\lVert\phi\rVert$ & $-0.011{\pm}0.002$ & $+0.023{\pm}0.004$ & $+0.032{\pm}0.001$ \\
    Cosine trained, raw $\phi^\top\psi$ & $+0.024{\pm}0.008$ & $-0.013{\pm}0.002$ & $-0.023{\pm}0.003$ \\
    Cosine trained, cosine readout & $+0.058{\pm}0.001$ & $-0.054{\pm}0.005$ & $-0.102{\pm}0.007$ \\
    \bottomrule
  \end{tabular}
\end{table}

Raw-trained embeddings lose some ordering after cosine normalization, though
their norms carry little ordering signal. Cosine-trained embeddings show weak
ordering under either readout.

\FloatBarrier

\section{Two-Head Value-Supervision Control}
\label[appendix]{app:twohead}

This control tests whether one checkpoint can support retrieval and a separate
value-supervised progress scalar. The network has a bounded cosine head
$f_{\cos}$ for retrieval and a Bellman/IQL head $q_{\mathrm{TD}}$ for
selection. In the \emph{separate} variant, $q_{\mathrm{TD}}$ has its own
encoder. In the \emph{joint} variant, both heads share the contrastive
backbone, which receives gradients from both objectives.

\Cref{tab:twohead} reports the dedicated two-head seed-42 evaluation; baseline
rows reuse the three-seed values from \cref{tab:main,tab:retrieval}. The
separate variant preserves retrieval while restoring graded progress ordering.
The shared-backbone model performs worse on HumanoidMaze. These results favor
using a separate value-supervised head over unrestricted parameter sharing. This
control does not test fixed-query candidate ordering.

\begin{table}[H]
  \centering
  \caption{\textbf{Retrieval and value-supervised progress ordering.}
    Cross-query $\gamma^d$ Kendall $\tau_b$ for the value head and retrieval
    AUC for the cosine head; two-head entries use dedicated checkpoints.}
  \label{tab:twohead}
  \begin{tabular}{l cc cc cc}
    \toprule
      & \multicolumn{2}{c}{PointMaze} & \multicolumn{2}{c}{AntMaze}
      & \multicolumn{2}{c}{HumanoidMaze} \\
    \cmidrule(lr){2-3}\cmidrule(lr){4-5}\cmidrule(lr){6-7}
    Critic & $\gamma^d\,\tau_b$ & AUC & $\gamma^d\,\tau_b$ & AUC
      & $\gamma^d\,\tau_b$ & AUC \\
    \midrule
    TD-Q baseline      & $+0.627$ & n/a     & $+0.598$ & n/a     & $+0.791$ & n/a     \\
    Cosine baseline    & $+0.058$ & $0.961$ & $-0.054$ & $0.989$ & $-0.102$ & $0.997$ \\
    Two-head, separate & $+0.635$ & $0.962$ & $+0.592$ & $0.991$ & $+0.785$ & $0.998$ \\
    Two-head, joint    & $+0.653$ & $0.963$ & $+0.557$ & $0.987$ & $+0.160$ & $0.998$ \\
    \bottomrule
  \end{tabular}
\end{table}